\documentclass{article}
\usepackage[utf8]{inputenc}

\usepackage{amsmath}
\usepackage{xcolor}

\usepackage{amssymb}
\usepackage{latexsym}
\usepackage{hyperref}

\usepackage{url}

\title{A Convolutional Architecture for 3D Model Embedding}
\author{Arniel Labrada, Benjamin Bustos, Ivan Sipiran}
\date{March 2021}

\usepackage{graphicx}

\begin{document}

\maketitle

\begin{abstract}
During the last years, many advances have been made in tasks like 3D model retrieval, 3D model classification, and 3D model segmentation. The typical 3D representations such as point clouds, voxels, and polygon meshes are mostly suitable for rendering purposes, while their use for cognitive processes (retrieval, classification, segmentation) is limited due to their high redundancy and complexity.
We propose a deep learning architecture to handle 3D models as an input. We combine this architecture with other standard architectures like Convolutional Neural Networks and autoencoders for computing 3D model embeddings. Our goal is to represent a 3D model as a vector with enough information to substitute the 3D model for high-level tasks. Since this vector is a learned representation which tries to capture the relevant information of a 3D model, we show that the embedding representation conveys semantic information that helps to deal with the similarity assessment of 3D objects. Our experiments show the benefit of computing the embeddings of a 3D model data set and use them for effective 3D Model Retrieval.
\end{abstract}

\section{Introduction}
Since the outstanding results obtained by AlexNet~\cite{DBLP:conf/nips/KrizhevskySH12} for image classification in 2012, the architectures of neural networks, specifically Convolutional Neural Networks (\emph{CNNs})~\cite{lecun1998gradient}, have been continuously improving to solve visual computing tasks. Especially in the image processing field, \emph{CNNs} have been used to solve the most relevant image processing tasks, outperforming the results obtained with the area's standard techniques. These tasks include retrieval, classification, segmentation, and computation of image embeddings, among others.

Due to this remarkable performance, researchers have extended the use of \emph{CNNs} to other fields such as 3D model processing. In recent years, some works propose using these networks to solve classification and retrieval of 3D models with excellent results~\cite{DBLP:journals/csur/IoannidouCNK17}. However, deep learning in the field of 3D models is still a relatively under-researched topic, and some 3D model tasks such as computing 3D model embeddings have received little to no attention.

In this work, we propose three different neural network architectures for computing 3D model embeddings. For this purpose, we first render a set of image views from the 3D models. To process these image view sets, we propose a neural network module to consume them as input. Then, we combine this module with other standard neural network architectures like \emph{CNNs} and autoencoders to obtain the 3D model embeddings.

Since 3D models are a complex and space-consuming type of data, their processing is often challenging. However, our proposal goal is to compile into single vectors as much information as possible of the 3D models. We can substitute the 3D models with these vectors in further tasks such as retrieval, cross-modal retrieval, classification, segmentation, and others. 
 
This paper presents the three main contributions:
\begin{itemize}
    \item We propose a convolutional architecture that can handle the 3D models represented as sets of image views.
    \item We give an analysis of the performance of different convolutional architectures for 3D model embedding.
    \item We study how the number of image views affects the quality of the 3D model embedding, by experimenting with different amount of image views (2, 3, 4) to represent the models.
\end{itemize}

Finally, we show how our proposed 3D shape embedding method can benefit 3D model retrieval. To achieve this goal, after we compute the vector representation of the 3D models, we use the cosine distance as the similarity metric between the vectors. With this methodology, we show that we can obtain outstanding results for the 3D model retrieval task.

\section{State of the Art}

Over the years, the 3D model research field has been highly relevant in Computer Science due to the steady increase in the number and use of 3D objects for many practical application domains. Surface segmentation and face recognition are some of the many important tasks in this field. However, two of the most addressed tasks are 3D shape retrieval and classification. Traditionally, the main approaches for fulfilling these tasks relied on hand-engineered feature extraction methods. Many kinds of features can be used for this purpose. Nevertheless, shape features are the most used because, in many cases,  one only has the shape of the 3D model as input data. Surveys in this area can be found in  Generic 3D shape retrieval~\cite{DBLP:conf/3dor/LiGABFLLJORTYZ12} and textured 3D model retrieval~\cite{DBLP:conf/3dor/CerriBAABCEFLGGLVVXZ13}.

\subsection{3D shape retrieval and classification using Neural networks}

Recently, some works propose using neural network techniques instead of extracting features to solve the 3D model retrieval and classification tasks with excellent results~\cite{DBLP:journals/csur/IoannidouCNK17}. This new approach shows promising results in the area. However, it is still a relatively under-researched topic, and some 3D model tasks such as computing 3D model embeddings have received little to no attention.

There are three main approaches for using deep learning techniques in the area of 3D shape retrieval and classification. These approaches are based on how 3D models are represented as input to a neural network model.

The first two approaches consist of representing 3D models as a set of image views and voxel grids, respectively~\cite{DBLP:journals/csur/IoannidouCNK17,DBLP:journals/corr/abs-1912-12033}. The representation of 3D models with image views yields better results than voxel grids in many applications. However, incorporating the voxel grid can lead to better results if one uses more complex neural network models, which are more expensive to train and need much more training data.

The third approach consists of the use of point cloud representations to train deep learning models. This approach has recently increased in popularity with some outstanding works like PointNet~\cite{DBLP:conf/cvpr/QiSMG17}, Dynamic Graph CNN for Learning on Point Clouds~\cite{DBLP:journals/corr/abs-1801-07829}, RS-CNN~\cite{DBLP:conf/cvpr/LiuFXP19}, and LDGCNN~\cite{DBLP:journals/corr/abs-1904-10014}. A survey on this topic can be found in Deep Learning for 3D Point Clouds: A Survey~\cite{DBLP:journals/corr/abs-1912-12033}.

\subsection{Image Views Representation}

Image views is a very feasible and accepted representation of 3D models for its processing. Since 3D models are a very complex and space costly type of data, in many cases, it is necessary to transform these objects into a more manageable kind of data such as an image or, as in this case, a set of images. Several works that use deep learning architectures in the 3D model field use this image view representation~\cite{DBLP:journals/csur/IoannidouCNK17}. The reason behind this is that training a deep learning model directly with any of the standards 3D models representation (voxel grid, polygon mesh) could be a very costly process both in time and resources. 

One of the earliest proposals for 3D shape retrieval using Convolutional Neural Networks is DeepEm~\cite{DBLP:journals/tip/GuoWGLL16}. They first propose an architecture for image embedding using a triple input for the training. The input consists of a query image, a positive image, and a negative image. Then, they compute the classification loss of each one using VGG19~\cite{DBLP:journals/corr/SimonyanZ14a} and a triplet loss using the last fully-connected layer of each of the three networks. Finally, to learn the embeddings, they jointly use all of the classification loss and the triplet loss. Using this network for image embedding, they also propose a 3D shape retrieval framework using Image Views representation. The goal is to compute embeddings from the image views and then use a set-to-set distance metric to calculate the similarity between two 3D shapes represented as a set of embeddings.

Su et al.~\cite{aktar2019multi} further explores the concept of using triplet loss with a convolutional neural network to compute image embedding and then using these embedding for 3D object retrieval. They also extend their proposal to unsupervised learning by using a convolutional autoencoder.

Su et al.~\cite{DBLP:conf/iccv/SuMKL15} proposed a standard \emph{CNN} trained to recognize the image views independently of each other, obtaining a very high accuracy for the recognition of 3D model even from a single view. Furthermore, they present a \emph{CNNs} architecture with an image view pooling layer that combines information from multiple views into a unique and compact shape descriptor, offering even better recognition performance.

This concept of using pooling for aggregations of view has been explored in some works with good results. Su et al.~\cite{DBLP:journals/cg/SfikasPT18} proposed an extension of the PANORAMA 3D shape representation~\cite{DBLP:journals/ijcv/PapadakisPTP10}. They use this representation as input to an ensemble of \emph{CNNs} with the goal of computing feature continuity of 3D models. They test their proposal for 3D model classification and retrieval against several other states of the art techniques achieving very competitive results.

More recently, Su et al.~\cite{DBLP:journals/tip/HanLLVLZHC19} further explore this approach of using pooling of views aggregation. The conjecture is that the redundant information within the views and their spatial relationships are lost during the pooling process. To solve this, they present a deep learning model (3D2SeqViews) with a novel hierarchical attention aggregation. This model not only aggregates the content information within all sequential views, but also the sequential spatiality among the views.

\subsection{Data Embedding}
One notably successful use of deep learning is obtaining data embedding representations. This technique aims to represent the data as vectors with enough information to substitute the data for different tasks. In the text processing field, several outstanding works apply this technique for words (e.g., Word2Vec~\cite{DBLP:journals/corr/Rong14}, Glove~\cite{DBLP:conf/emnlp/PenningtonSM14}) and sentence embedding (e.g., InferSent~\cite{DBLP:conf/emnlp/ConneauKSBB17}, BERT~\cite{DBLP:conf/naacl/DevlinCLT19}). 

There is a plethora of research available on image embedding, as well. Many researchers have proposed new architectures of \emph{CNNs} to compute embedding for images. Examples of these works are LIFT~\cite{DBLP:conf/eccv/YiTLF16}, ``Learning Image Embeddings using Convolutional Neural Networks for Improved Multi-Modal Semantics''~\cite{DBLP:conf/emnlp/KielaB14}, and ``Deep Image Retrieval: Learning Global Representations for Image Searc''~\cite{DBLP:conf/eccv/GordoARL16}. However, there is not much research on computing 3D shape embedding. Moreover, the few works that compute embedding for 3D shapes are mainly based on point cloud representations~\cite{DBLP:conf/iclr/AchlioptasDMG18,DBLP:conf/iclr/ChenCM20} or voxel grids. We can also find a proposal for computing 3D shapes embedding using engineered features and the bag-of-words framework~\cite{DBLP:journals/access/LiSDZCD18}. However, although some proposals indirectly use the concept, to our knowledge, no method has been proposed yet for 3D model embeddings using image view representation and \emph{CNNs}.

\section{Background Knowledge}

In this section, we discuss some background concepts for a better understanding of our proposal.

\subsection{Convolutional Neural Networks (\emph{CNNs})}

Convolutional Neural Networks (\emph{CNNs})~\cite{lecun1998gradient} are models of neural networks that have shown outstanding performance in different tasks. In the area of image processing, since the excellent results obtained in 2012 by AlexNet~\cite{DBLP:conf/nips/KrizhevskySH12}, these networks have been used to solve various tasks like image recognition, image classification, object detection, and face recognition. 

The standard architecture of \emph{CNNs} consists of a series of convolutional layers and pooling layers. The convolutional layers employ a convolution operation instead of simple matrix multiplication applied in traditional neural networks. The output of a convolutional layer is a feature map as the result of the convolution operation with a defined $n \times n$ kernel. Thus, the network learns the filters that in traditional algorithms were hand-engineered.  On the other hand, the pooling layers are used to reduce the dimensionality of the feature maps. There are several pooling layers, for example, max pooling, average pooling, and sum pooling.

For the image classification task, we use one or more fully connected layers on top of the convolutional layers and pooling layers. Then, we connect the last fully connected layer to a classifier that is usually a softmax function. Finally,  we use the cross-entropy as the loss function.

\subsection{Autoencoders}

\begin{figure*}[ht]
\begin{center}
  \includegraphics[width=\linewidth]{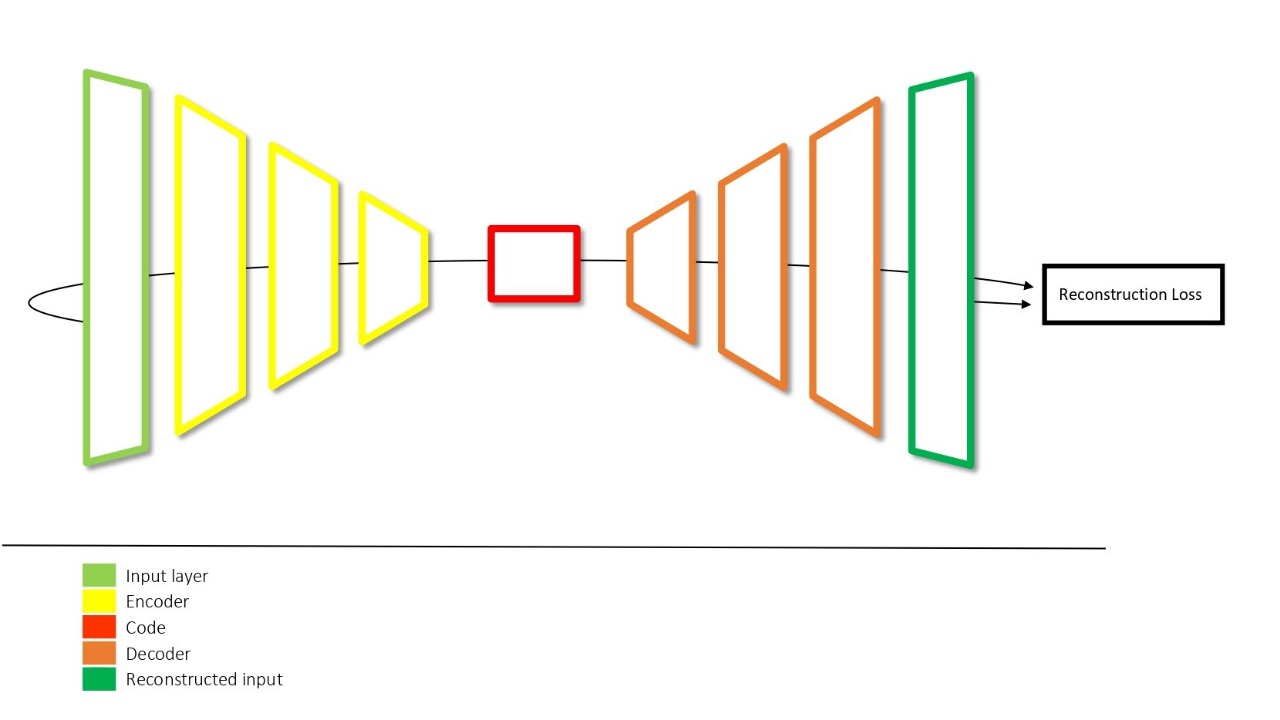}
  \caption{\label{fig1}A standard autoencoder architecture}
 \end{center}
\end{figure*}

Autoencoders~\cite{DBLP:conf/aaai/Ballard87} are an unsupervised learning technique mainly used for data representation (embedding, deep features, etc.) and data compression. We create these networks to impose a bottleneck in their design, which forces a compressed knowledge representation of the original input. In more detail, an autoencoder architecture consists of four parts: encoder, bottleneck, decoder, and reconstruction loss. The encoder learns how to reduce the input dimensions; the bottleneck is the layer that contains the compressed representation of the input data; the decoder reconstructs the data from the bottleneck; the reconstruction loss measures how well the decoder is performing and how close the output is to the original input. Figure~\ref{fig1} shows a standard autoencoder architecture.

In the image processing field, these architectures are used as well for image compression and image representation. The most common network used to this end is the convolutional autoencoder, which uses the mentioned encoder-decoder architecture with some adjustment. The encoder part of a convolutional autoencoder consists of a standard convolutional neural network built with convolutional layers and pooling layers to reduce the resolution. Then, the encoder is connected to a fully connected layer, which serves as the bottleneck. Finally, the bottleneck is the input to the network's decoder, built with what is known as de-convolutional layers and un-pooling layers. A de-convolutional layer is just the transposed of its corresponding convolutional layer. For the un-pooling layers, it is a bit more complicated. These layers are used to increase the resolution, the inverse operation of its corresponding pooling layer. The problem is that the usual pooling layers used, like max-pooling, are non-invertible operations. For this reason, several un-pooling techniques have been defined in the last years~\cite{DBLP:journals/pami/ChenPKMY18,DBLP:journals/corr/TurchenkoCL17}.

\section{A New Convolutional Architecture for 3D Model Embedding and Retrieval}

We propose a convolutional architecture for 3D model embedding. The idea is to obtain a 3D model deep representation for effective 3D model retrieval. To this end, first, we represent the 3D models as a set of image views. Then, this representation is used as input to train our proposed convolutional model. The problem here is that a standard convolutional network uses an image as input, and in our case, we have a set of images as input. To solve this, we model the information as a multi-channel image where the number of channels is the number of image views that we use to represent a 3D model. With our proposed architecture to handle the image view sets as input, we develop three convolutional architectures for 3D model embedding.

\subsection{Autoencoder Network}
 Our first proposal is a convolutional autoencoder architecture. In this model, we connect our proposed multi-channel input to a standard convolutional neural network, which serves as the model's encoder. Then, we connect the encoder to a fully connected layer (the bottleneck), which, at the same time, we link to the decoder of the network.

The network's encoder has four blocks of two convolutional layers and one max pooling layer with stride two, which means that each block reduces the dimensionality by half. Each of the convolutional layers has a kernel of $5 \times 5$ and channels equal to $64*2^{k}$, where $k$ is the respective block $(1,2,3,4)$.

\begin{figure*}[!t]
  \includegraphics[width=\linewidth]{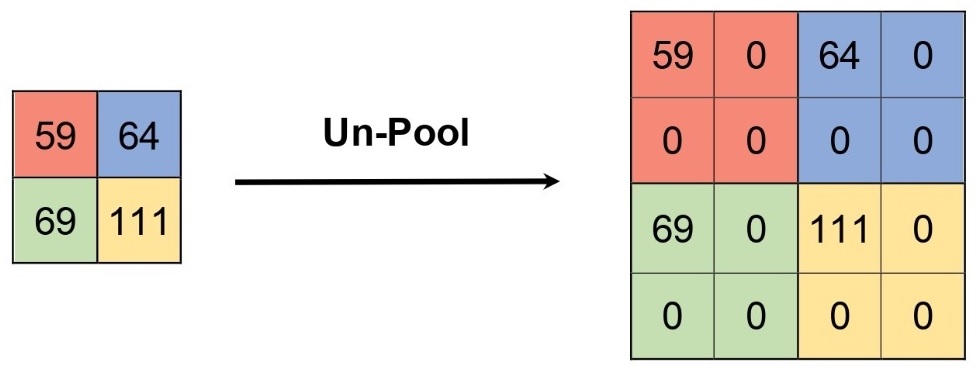}
  \caption{Un-pooling layer used for the decoder of the model}
  \label{fig6}
\end{figure*}

The network's decoder also has four blocks composed of one transposed convolutional layer, also known as a deconvolutional layer. One un-pooling layer with stride two doubles the dimensionality of each block. Figure~\ref{fig6} shows the un-pooling layer used in the model. Each of the deconvolutional layers has a kernel of $5 \times 5$ and channels equal to $64*2^{k}$, where $k$ is the respective block $(4,3,2,1)$.

\begin{figure*}[!t]
  \includegraphics[width=\linewidth]{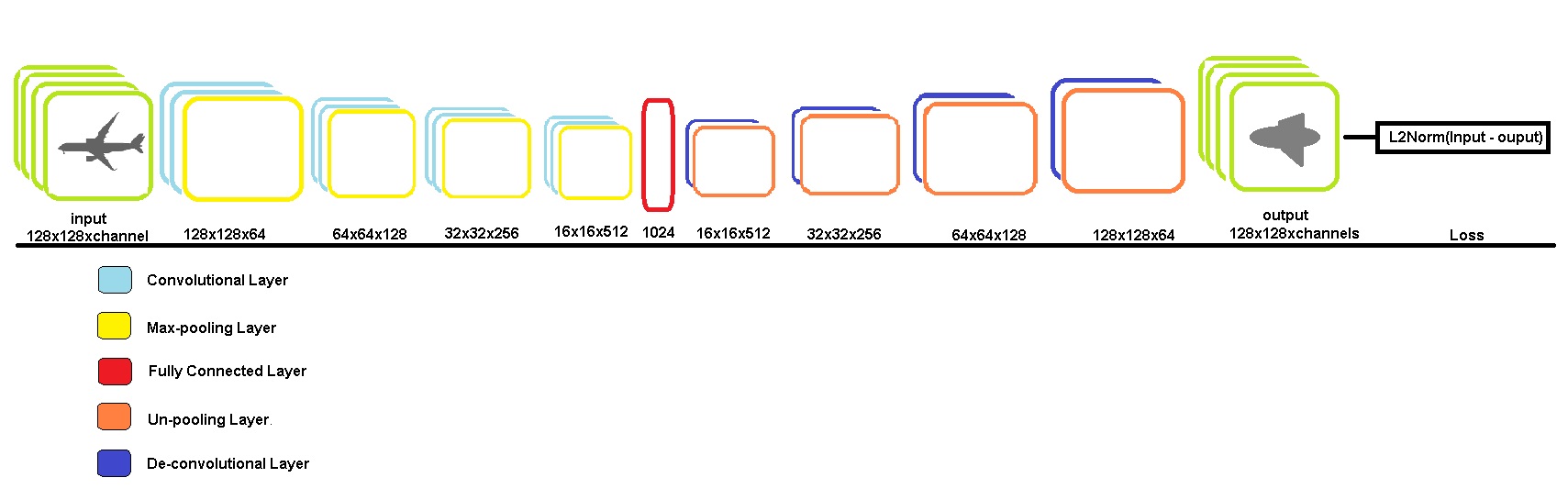}
  \caption{Convolutional autoencoder for 3D model embedding}
  \label{fig2}
\end{figure*}

Finally, we use as reconstruction loss the $L_{2}$ norm of the difference between the output and the model's input. Figure~\ref{fig2} shows our convolutional autoencoder model.

\subsection{Classification Network}

\begin{figure*}[!t]
  \includegraphics[width=\linewidth]{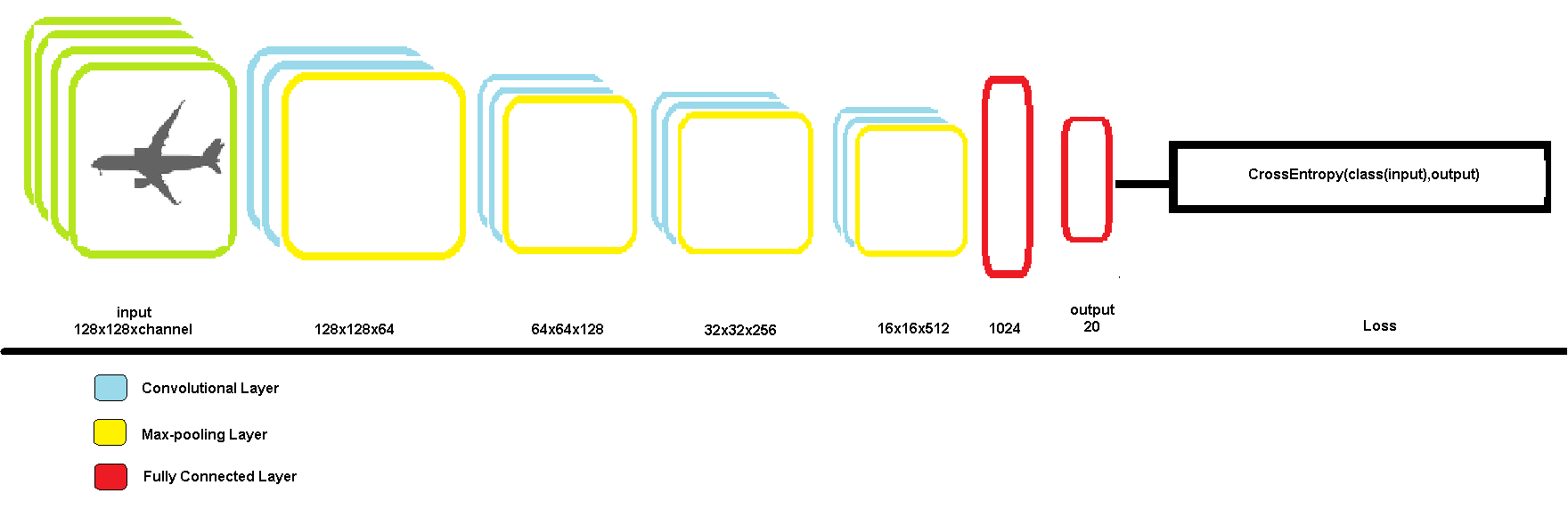}
  \caption{Classification network for 3D model embedding}
  \label{fig3}
\end{figure*}

Autoencoder models are dominant in compressing and computing representations of the data. However, these models do not learn relations between objects because they only use the input information to train. We address this problem with our second proposal, which consists of a \emph{CNN} for classification. This proposed network uses the same architecture as the autoencoder proposal until the bottleneck component. After that, we connect the bottleneck to a fully connected layer of size $C$, where $C$ is the number of classes for classification. Then, we apply a standard soft-max activation function over this layer. Finally, we use cross-entropy as the loss function. We train this model until we obtain very high accuracy, and then, we use the bottleneck layer as the embedding. Figure~\ref{fig3} shows our convolutional model for classification.

\subsection{Combination of the autoencoder and the classification netwok}

\begin{figure*}[!t]
  \includegraphics[width=\linewidth]{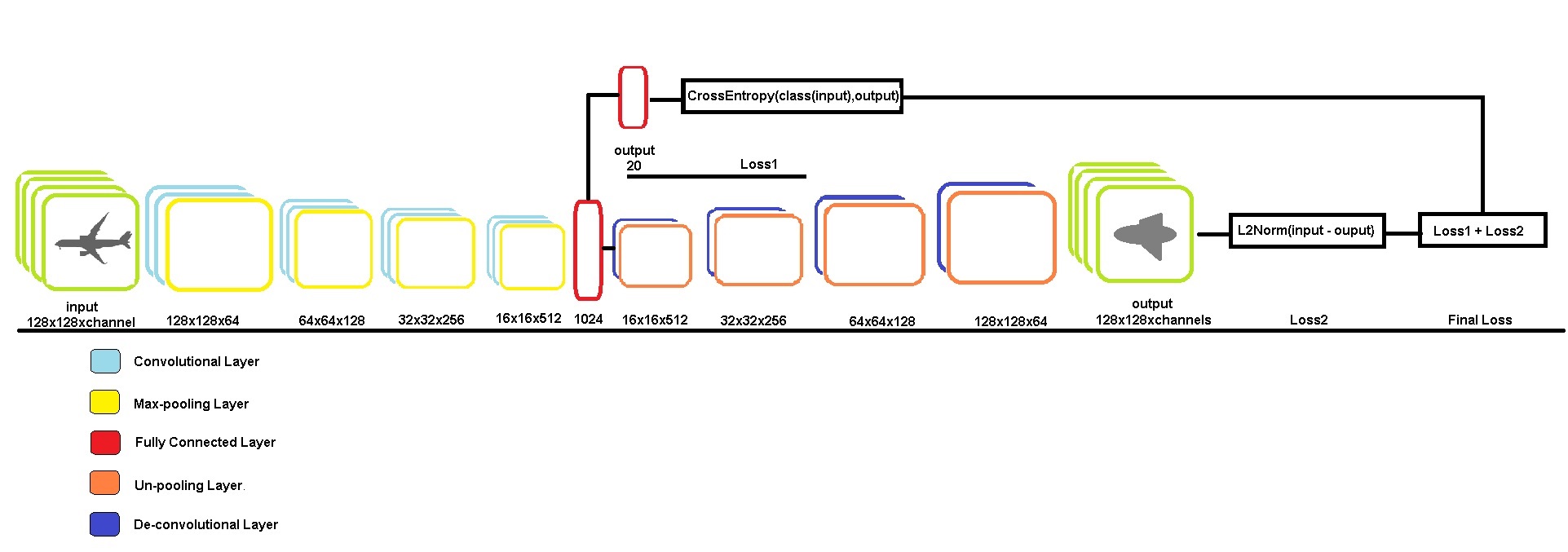}
  \caption{Combination of convolutional autoencoder and classification network for 3D model embedding}
  \label{fig4}
\end{figure*}

Finally, we want our 3D model embedding network to have high compression power and learn the relation between the 3D models. With that in mind, our last proposal consists of a combination of both already described models. This new model uses the autoencoder and the classification network architecture at the same time. In other words, for a given input, the model computes the loss of the autoencoder and the loss of the classification network, and the final loss of this new network is the sum of both already calculated losses. After the training, we use once again the bottleneck layer as the embedding of the 3D models. Figure~\ref{fig4} shows this last proposed model.

\section{Evaluation Metrics}

We use two standard performance metrics for retrieval techniques to evaluate our proposal for 3D model embedding applied to 3D model retrieval. These metrics are based on a ranked list, consisting of all 3D models ordered according to their dissimilarity with a specific query. The metrics are Mean Average Precision \emph{MAP}, and Discounted Cumulative Gain \emph{DCG}. Their meanings and definitions are explained below.

The Mean Average Precision \emph{MAP} of a set of \textit{N} queries is calculated as follows: for each query, \textit{q}, we compute its corresponding Average Precision \emph{AP}, and then, the mean of all these scores (Eq. \ref{MAP}). The resultant value measures the quality of models at carrying out queries and is approximately equal to the area under the precision-recall curve \cite{zuva2012evaluation}.

\begin{equation}
\label{MAP}
    MAP =  \frac{\sum_{q=1}^{N} AP(q)}{N}
\end{equation}

The second metric, the \emph{DCG} metric, uses the fact that users of a search engine are more interested in the top results of the queries they execute. For this reason, to compute the final score, \emph{DCG} uses all the elements of the ranked list that belong to the same class as the query. However, the closer to the beginning of the ranked list the elements appear, the more weight they provide to the final score. Thus, the metric follows three steps. First, the ranked list $R$ is converted to a list $G$, where element $G_i$ has value $1$ if element $R_i$ is in $C$ and value $0$ otherwise, where $C$ is the class of the query image. Second, the discounted cumulative gain is calculated as follows:

\begin{equation}
DCG_i = \left\{\begin{matrix}
G_{1} & i = 1\\ 
DCG_{i-1} + G_i/log & i \neq 1
\end{matrix}\right.
\end{equation}

\noindent Finally, the obtained result is divided by the maximum possible \emph{DCG}:

\begin{equation}
DCG = \frac{DCG_n}{1 + \sum_{j = 2}^{\left |C  \right |} \frac{1}{\log_2 j}}
\end{equation}

\noindent where $n$ is the size of $R$.

\subsection{Micro and macro average}

Along with these two evaluation metrics, MAP and DCG, we also use two versions of each metric. These versions are the micro averaged and the macro averaged.
On the one hand, the micro averaged version will aggregate the contributions of every object to calculate the average metric. On the other hand, the macro averaged version will compute the metric for each class of objects and average these results. In multiclass classification and retrieval, the micro averaged version is more desirable if one knows that there is an imbalance in the categories' sizes.

\section{Experiments and Results}

In this section, we discuss all the experimental setups,describe the benchmark and the preprocessing of the data, and give the technical aspects of our three proposed architectures.

\subsection{Data configurations}

We use the ShapeNet~\cite{DBLP:journals/corr/ChangFGHHLSSSSX15} benchmark dataset for the experiments. This dataset comprises 51,300 3D models grouped into 55 categories, and they are provided in OBJ format. The dataset count with two versions, consistently aligned(normalized dataset), and a more challenging dataset where random rotations perturb models. The dataset provides a split 70\%/10\%/20\% for training, validation, and test, respectively. We conduct several experiments using both versions of the dataset.

As mentioned before, our proposed models consume sets of image views as input. An image view is a picture of a 3D model from a predefined viewpoint. So, after choosing our dataset of 3D models for the experiment, we have to render the image views from each 3D model. In our case, we first extract 30 image views per 3D model.

To extract the image views, we use the Stanford-Shapenet-renderer script. This script uses the API from the Blender software to render images from different angles of a given 3D model in OBJ format. This code is public on Github~\footnote{\href{https://github.com/panmari/stanford-shapenet-renderer}{\bf Stanford-Shapenet-renderer}}
 and fully parameterizable for different tasks. In our case, we use the code to render images from the 3D models starting with the camera, pointing to the coordinates origin. From there, we render an image every 12 degrees in circumference around the 3D model itself.

However, our models are not powerful enough to handle sets of 30 image views as input. To solve this problem, we add a new preprocessing step to the image view representation to reduce considerably the number of image views used to represent the 3D models. This new step consists of using a clustering algorithm over the set image views representing a 3D model to group them in $k$ different clusters. After that, we choose a representative from each cluster, and the final representation of the 3D model is the set of all $k$ representatives. In our case, we use the k-means algorithm for the clustering of the image views with $k$ centroids. As for the representatives, we choose the image view most similar to each centroid. To study the impact of the number of image views representing the 3D models, we use different parameter $k$. Specifically, we choose $k$ in the set (2,3,4).

\subsection{Proposal setups and results}

We propose three different \emph{CNNs} architectures for computing 3D model embeddings. For all three proposals, we use \textit{adam} for the optimization, \textit{relu} as the activation function, and a batch of size 100 for the training. In the case of the Autoencoder and Combination models (first and third proposed model, respectively), we train for 50,000 iterations since the autoencoder component takes more time to converge. On the other hand, for the Classification model (our second proposal), we only use 20,000 iterations since the classification accuracy quickly achieves values above 98 percent.

Our goal is to use the embeddings computed with our proposals for 3D model retrieval. To do that, we use the cosine distance to measure the similarity between the embeddings and build a ranked list for each 3D model. After that, we apply the two mentioned metrics to measure retrieval performance over the ranked lists. Table~\ref{table1} shows the micro averaged results of our proposed models using the normalized version of the dataset. We can see that the autoencoder model is the worst, and the combination of the autoencoder with the classification network achieves the best. We can also appreciate that the number of image views affects the result's quality: the more image views, the best performance of each model.

We also compare our proposed model against the four proposals that reported the best results in the ShapeNet dataset (RotationNet, GIFT, ReVGG, DLAN)~\cite{zuva2012evaluation} using both versions of the dataset. Table~\ref{table2} and Table~\ref{table3} show these comparisons using the normalized and the perturbed version of the dataset. We also compute the micro and macro averages of both metrics used in our experiments.

The results show that our proposed model achieves very competitive effectiveness, specially in the normalized version of the dataset where we obtain a very high \emph{DCG}. We note that the embedding quality is affected by the number of image views representing the 3D models.
For the three proposed methods, as the number of image views increases, the performance also increases. However, using a different number of image views makes little change in our last method's performance in the normalized version of the dataset.


We also note that the effectiveness of the combined method is affected  when we use the perturbed version of the dataset. Nevertheless, when we increase the number of image views per 3D model in the dataset's perturbed version, the effectiveness increases notoriously. This phenomenon indicates that we need more images to capture each 3D model's information in the perturbed version. If each object has an arbitrary orientation, the chances of obtaining similar views in different objects is proportional to the number of extracted views. However, increasing the number of image views representing the 3D model could be very costly for our proposed model since our model's autoencoder component is computationally expensive to train. Adding more layers and channels could lead to a better performance at the cost of increasing the computational load. An altenative solution is the pose normalization of the objects in the dataset.


Another important observation in our experiment results is that the two evaluation metrics, the \emph{DCG} and the \emph{MAP}, are more distant than in most other authors' results. It happens because the relevant result obtained for a given query using our proposed model are accumulated at the beginning and the end of the ranked list. This result is a useful feature for a search engine that we can exploit in future works using our proposed model.

For reproducibility purposes, we made available the code
\footnote{Anonymized.}
of the three proposed embedding models.
\begin{table*}[!t]
\caption{Evaluation of the performance of the embedding for 3D models Retrieval using the normalized version of the ShapeNet.}
\label{table1}
\centering
\begin{tabular}{|p{3cm}|c|p{2cm}|p{2cm}|}
\hline
\multicolumn{2}{|c}{Metric Calculation} & \multicolumn{2}{|c|}{Micro Average} \\
\hline
Model          & Views & DCG    & MAP    \\ \hline
Autoencoder    & 2     & 0.649 & 0.276 \\ \hline
Autoencoder    & 3     & 0.655 & 0.288 \\ \hline
Autoencoder    & 4     & 0.655 & 0.290 \\ \hline
Classification & 2     & 0.842 & 0.501 \\ \hline
Classification & 3     & 0.846 & 0.527 \\ \hline
Classification & 4     & 0.857 & 0.567 \\ \hline
Autoe.+Class.  & 2     & 0.895 & 0.663  \\ \hline
Autoe.+Class.  & 3     & 0.897 & 0.679 \\ \hline
Autoe.+Class.  & 4     & \textbf{0.901} & \textbf{0.684} \\ \hline
\end{tabular}
\end{table*}

\begin{table*}[!t]
\caption{Comparison of our computed embedding against other 3D model retrieval methods using the normalized version of the ShapeNet.}
\label{table2}
\centering
\begin{tabular}{|p{3cm}|c|p{2cm}|p{2cm}|p{2cm}|p{2cm}|}
\hline
\multicolumn{2}{|c}{Metric Calculation} & \multicolumn{2}{|c}{Micro Average} & \multicolumn{2}{|c|}{Macro Average} \\
\hline
Model          & Views & DCG    & MAP & DCG    & MAP   \\ \hline
RotationNet    & -     & 0.865 & \textbf{0.772}  & 0.656 & \textbf{0.583} \\ \hline
GIFT            & -     & 0.827 & \textbf{0.722} & 0.657 & 0.0575 \\ \hline
ReVGG           & -     & 0.828 & 0.749 & 0.559 & 0.496 \\ \hline
DLAN            & -     & 0.762 & 0.663 & 0.563 & 0.477\\ \hline
Autoe.+Class.  & 2     & 0.895 & 0.663  & 0.754 & 0.453\\ \hline
Autoe.+Class.  & 3     & 0.897 & 0.679 & 0.765 & 0.461\\ \hline
Autoe.+Class.  & 4     & \textbf{0.901} & 0.684 & \textbf{0.768} & 0.466\\ \hline
\end{tabular}
\end{table*}

\begin{table*}[!t]
\caption{Comparison of our computed embedding against other 3D model retrieval methods using the perturbed version of the ShapeNet.}
\label{table3}
\centering
\begin{tabular}{|p{3cm}|c|p{2cm}|p{2cm}|p{2cm}|p{2cm}|}
\hline
\multicolumn{2}{|c}{Metric Calculation} & \multicolumn{2}{|c}{Micro Average} & \multicolumn{2}{|c|}{Macro Average} \\
\hline
Model          & Views & DCG    & MAP & DCG    & MAP   \\ \hline
RotationNet    & -     & 0.702 & 0.606  & 0.407 & 0.327 \\ \hline
GIFT            & -     & 0.701 & 0.567 & 0.513 & 0.406 \\ \hline
ReVGG           & -     & \textbf{0.783} & \textbf{0.696} & 0.479 & 0.418 \\ \hline
DLAN            & -     & 0.754 & 0.656 & \textbf{0.560} & \textbf{0.476}\\ \hline
Autoe.+Class.  & 2     & 0.730 & 0.298  & 0.453 & 0.141\\ \hline
Autoe.+Class.  & 3     & 0.743 & 0.320 & 0.467 & 0.160\\ \hline
Autoe.+Class.  & 4     & 0.756 & 0.346 & 0.476 & 0.182\\ \hline
\end{tabular}
\end{table*}

\section{Conclusions}


In this work, we propose three different neural network architectures for computing 3D model embeddings. The first one is based on convolutional autoencoders. The second one is based on a convolutional neural network for classification. The last one is a combination of the first two. For all of our three proposals, we use the same technique for modeling the input. This technique involves transforming a set of image views representing a 3D model into a multichannel image where each channel is an image view. We conduct several experiments using our proposals, and we can conclude that our work has three main contributions. A convolutional architecture that can handle the 3D models represented as sets of image views, an analysis of the performance of different convolutional architectures for computing 3D model embedding, and a study of how the number of image views affects the quality of the computed embedding by experimenting with different amount of image views $(2, 3, 4)$ to represent the models.

The experiments show that the model with the worse results was the autoencoder since, as we already mentioned, this model does not learn relations between the 3D models. Instead, to compute the embeddings for a given 3D model, it uses the 3D model itself. On the other hand, our second proposal highly outperforms the first one. Since the second proposal uses a classification network, this model intrinsically learns relations between the 3D models, specifically, if two 3D models belong to the same class or not. However, we obtain the best result using our last proposal, which combines the first two models.

The most important conclusion is that our proposed models can successfully compute embedding representations for 3D models. We obtain outstanding results for the two evaluation metrics, especially with our last model, which shows results for the \emph{MAP} above 67 percent and \emph{DCG} above 90 percent. These results are very competitive with the Shapenet Dataset results, especially in the normalized version of the dataset.

\end{document}